\documentclass[letterpaper, 10 pt, conference]{ieeeconf}
\IEEEoverridecommandlockouts   % if \thanks is used
\usepackage{xcolor}
\usepackage{cite}
\usepackage{graphicx}
\usepackage{epsfig} % for postscript graphics files
\usepackage{amssymb}  % assumes amsmath package installed
\usepackage{url}            % simple URL typesetting
\usepackage{amsmath} % assumes amsmath package installed
\usepackage{microtype}      % microtypography
\usepackage{color}
\usepackage{bm}
\usepackage[detect-all]{siunitx}
\usepackage{hyperref}
\usepackage[capitalise]{cleveref}
\usepackage{balance}
\usepackage{cancel}
\usepackage{diagbox}
\usepackage{hyperref}
\graphicspath{{./figures/}}

\newcommand{\B}[1]{\mathbf{#1}}
\newcommand{\ns}{\textit{Neural-Swarm}}
\newtheorem{assumption}{Assumption}
\newtheorem{theorem}{Theorem}   %[section]

\newtheorem{lemma}[theorem]{Lemma}

\crefname{lemma}{lemma}{lemmas}
\Crefname{lemma}{Lemma}{Lemmas}
\crefname{thm}{theorem}{theorems}
\Crefname{thm}{Theorem}{Theorems}
\crefname{section}{Sec.}{sections}
\crefname{figure}{Fig.}{Fig.}
\crefformat{equation}{(#2#1#3)}
\crefrangeformat{equation}{(#3#1#4) to~(#5#2#6)}
\crefmultiformat{equation}{(#2#1#3)}%
{ and~(#2#1#3)}{, (#2#1#3)}{ and~(#2#1#3)}
\crefrangemultiformat{equation}{(#3#1#4) to~(#5#2#6)}%
{ and~(#3#1#4) to~(#5#2#6)}{, (#3#1#4) to~(#5#2#6)}{ and~(#3#1#4) to~(#5#2#6)}

\title{\LARGE \bf
Neural-Swarm: Decentralized Close-Proximity Multirotor Control\\ Using Learned Interactions
}

\author{Guanya Shi, Wolfgang  H\"onig, Yisong Yue, and Soon-Jo Chung
\thanks{Demo videos: \url{https://youtu.be/v4j-9pH11Q8}}
\thanks{The authors are with California Institute of Technology, USA.
\texttt{\{gshi, whoenig, yyue, sjchung\}@caltech.edu}.}
\thanks{We thank Michael O'Connell and Xichen Shi for helpful discussions and Anya Vinogradsky for help with the firmware implementation.}
\thanks{The work is funded in part by Caltech's Center for Autonomous Systems and Technologies (CAST) and the Raytheon Company.}
}

\begin{document}

\maketitle
\thispagestyle{empty}
\pagestyle{empty}

%%%%%%%%%%%%%%%%%%%%%%%%%%%%%%%%%%%%%%%%%%%%%%%%%%%%
\begin{abstract}
In this paper, we present \ns{}, a nonlinear decentralized stable controller for close-proximity flight of multirotor swarms. Close-proximity control is challenging due to the complex aerodynamic interaction effects between multirotors, such as downwash from higher vehicles to lower ones. Conventional methods often fail to properly capture these interaction effects, resulting in controllers that must maintain large safety distances between vehicles, and thus are not capable of close-proximity flight. Our approach combines a nominal dynamics model with a regularized permutation-invariant Deep Neural Network (DNN) that accurately learns the high-order multi-vehicle interactions. We design a stable nonlinear tracking controller using the learned model. Experimental results demonstrate that the proposed controller significantly outperforms a baseline nonlinear tracking controller with up to four times smaller worst-case height tracking errors. We also empirically demonstrate the ability of our learned model to generalize to larger swarm sizes. 
\end{abstract}

\section{Introduction}
The ongoing commoditization of unmanned aerial vehicles (UAVs) is propelling interest in advanced control methods for large aerial swarms~\cite{swarmsurvey,morgan2016swarm}. Potential applications are plentiful, including 
 manipulation, search, surveillance,  mapping, amongst many others.
Many settings require the UAVs to fly in close proximity to each other, also known as dense formation control.
For example, consider a search-and-rescue mission where the aerial swarm must enter and search a collapsed building. In such scenarios,
close-proximity flight enables the swarm to navigate the building much faster compared to swarms that must maintain large distances from each other.

A major challenge of close-proximity control is that the small distance between UAVs creates complex aerodynamic interactions.
For instance, one multirotor flying above another causes the so-called downwash effect on the lower one, which is difficult to model using conventional approaches~\cite{kumardownwash}.
%This interaction is known as downwash and can be included as constraint during motion planning.
In lieu of better downwash interaction modeling, one must require a large safety distance between vehicles, e.g., \SI{60}{cm} for the small Crazyflie 2.0 quadrotor (\SI{9}{cm} rotor-to-rotor)~\cite{DBLP:journals/trob/HonigPKSA18}.
However, the downwash for two Crazyflie quadrotors hovering \SI{30}{cm} on top of each other is only \SI{-9}{g}, which is well within their thrust capabilities, and suggests that proper modeling of downwash and other interaction effects can lead to more precise dense formation control.

In this paper, we propose a learning-based controller, \ns, to improve the precision of close-proximity control of homogeneous multirotor swarms.  
%take a learning approach to model the interaction forces for a homogeneous multi-quadrotor swarm. 
In particular, we train a regularized permutation-invariant deep neural network (DNN) to predict the residual interaction forces not captured by nominal models of free-space aerodynamics.  
The DNN only requires relative positions and velocities of neighboring multirotors as inputs, similar to existing collision-avoidance techniques~\cite{DBLP:conf/isrr/BergGLM09}, which enables a fully decentralized computation.
We use the predicted interaction forces as a feed-forward term in the multirotors' position controller, which enables close-proximity flight.
Our solution is computationally efficient and can run in real-time on a small 32-bit microcontroller.
We validate our approach on different tasks using two to five quadrotors.
To our knowledge, our approach is the first that models interactions between more than two multirotor vehicles.

\begin{figure}[t]
\includegraphics[width=\linewidth]{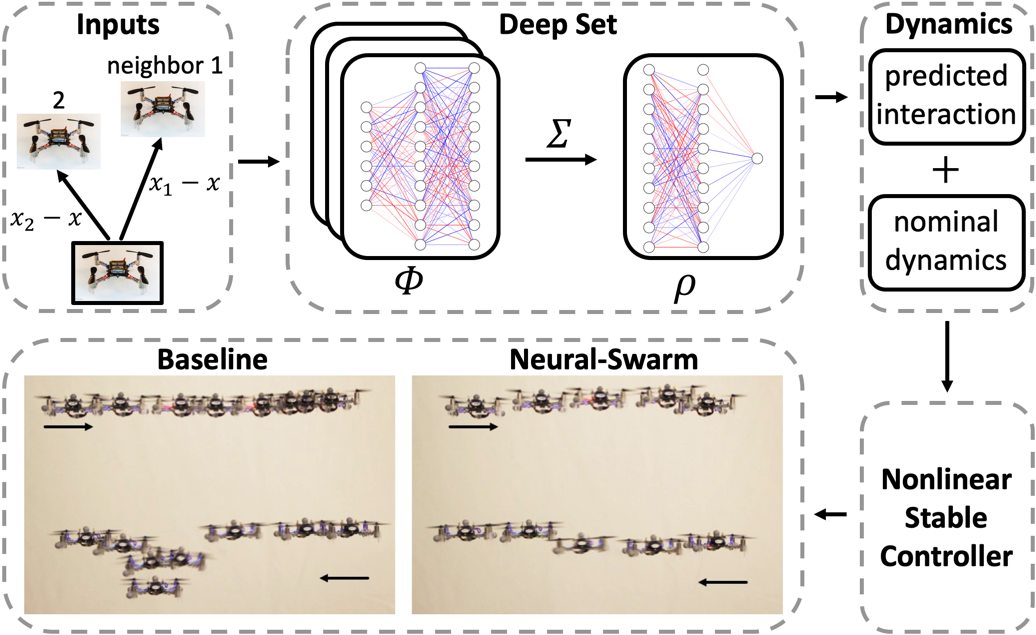}
% \caption{Close-proximity flight of two quadrotors swapping sides. Left: baseline controller which cannot maintain formation due to downwash; Right: \ns{} which properly accounts for downwash.}
\caption{We learn complex interaction between multirotors using regularized deep sets and design nonlinear stable controller for close-proximity flight.}
\label{fig:intro}
\vspace{-15pt}
\end{figure}

From a learning perspective, we leverage two state-of-the-art tools to arrive at effective DNN models.  The first is spectral normalization~\cite{bartlett2017spectrally}, which ensures the DNN is Lipschitz continuous.  As in our prior work~\cite{shi2019neural-lander}, Lipschitz continuity enables us to derive stability guarantees, and also helps the DNN generalize well on test examples that lie outside the training set.  We also employ deep sets~\cite{deepsets} to encode multi-vehicle interactions in an index-free or permutation-invariant manner, enabling better generalization to new formations and varying number of vehicles.

% Aerial swarm is an area of intense research within the robotics and control community, and the use of aerial swarm to solve real world problems has been increasing steadily~\cite{swarmsurvey,crazyswarm}.

% Aerial swarm requires high precision control. For a dense swarm case, this problem becomes challenging due to the complex aerodynamical interactions between vehicles. For example, consider two hovering 31 gram small quadrotors with vertical distance 0.5m, 

% In this paper,

%\todo{motivation: understand robot-robot aerodynamics forces to a) fly closer in formation, b) motion planning}
%\todo{emphasize that distributed algorithms that only relies on relative positions is important (vs centralized algorithms that requires full state of all other robots)}
%\todo{emphasize that we only consider the homogeneous case}

\subsubsection*{Related Work}
The use of DNNs to learn higher-order residual dynamics or control outputs is becoming increasingly common across a range of control and reinforcement learning settings \cite{shi2019neural-lander,le2016smooth,taylor2019episodic,cheng2019control,mckinnon2019learn,saveriano2017data,johannink2018residual}. The closest approach to ours is the Neural Lander~\cite{shi2019neural-lander}, which uses a DNN to capture the interaction between a single UAV and the ground, i.e., the well-studied ground effect \cite{cheeseman1955effect, doi:10.2514/6.2015-1685,kumardownwash}.  In contrast, our work focuses on learning inter-vehicle aerodynamic interactions between several multirotors.

%A well studied interaction between a single UAV and the environment is the ground effect.
%Various mathematical models have been proposed to estimate the magnitude of the ground effect~\cite{cheeseman1955effect, doi:10.2514/6.2015-1685,kumardownwash}.
%Moreover, this effect can be estimated using our prior work, a spectrally normalized neural network~\cite{shi2019neural-lander}.
%In contrast, we focus on inter-robot interactions using learning.

The interaction between two rotor blades has been studied in a lab setting to optimize the placement of rotors on a multirotor~\cite{drones2040043}.
However, it remains an open question how this influences the flight of two or more multirotors in close proximity.
Interactions between two multirotors can be estimated using a propeller velocity field model~\cite{twoagent}.
Unfortunately, this method is hard to generalize to the multi-robot case and this method only considers the stationary case, which will not work for many scenarios like swapping in \cref{fig:intro}.
We instead use a learning-based method that can directly estimate the interaction forces of multiple neighboring robots from training data.

For motion planning, empirical models have been used to avoid harmful interactions~\cite{morgan2016swarm,morgan2014model,DBLP:conf/iros/DebordHA18, DBLP:conf/icra/MellingerKK12}.
Typical safe interaction shapes are ellipsoids or cylinders and such models work for homogeneous and heterogeneous multirotor teams.
Estimating such shapes requires potentially dangerous flight tests and the shapes are in general conservative.
In contrast, we use learning to estimate the interaction forces accurately and use those forces in the controller to improve trajectory tracking performance in close-proximity flight.
The learned forces can potentially be used for motion planning as well.

%\todo{add newest vijay kumar paper here}

\section{Problem Statement: Swarm Interactions}
\subsection{Single Multirotor Dynamics}
A single multirotor's state comprises of the global position $\B{p} \in \mathbb{R}^3$, global velocity $\B{v} \in \mathbb{R}^3$, attitude rotation matrix $R \in \mathrm{SO}(3)$, and body angular velocity $\bm{\omega} \in \mathbb{R}^3$.
We consider the following dynamics:
\begin{subequations}
\begin{align}
\dot{\B{p}} &= \B{v}, &  
m\dot{\B{v}} &=m\B{g}+R\B{f}_u + \B{f}_a,\label{eq:pos_dynamics} \\ 
\dot{R}&=RS(\bm{\omega}), & 
J\dot{\bm{\omega}} &= J \bm{\omega} \times \bm{\omega}  + \bm{\tau}_u + \bm{\tau}_a,
\label{eq:att_dynamics}
\end{align}
\label{eq:dynamics}
\end{subequations}
\hspace{-5pt}where $m$ and $J$ are the mass and inertia matrix of the system, respectively; $S(\cdot)$ is a skew-symmetric mapping; $\B{g} = [0, 0, -g]^\top$ is the gravity vector; and $\B{f}_u = [0, 0, T]^\top$ and $\bm{\tau}_u = [\tau_x, \tau_y, \tau_z]^\top$ are the total thrust and body torques from the rotors, respectively. The output wrench $\bm{\eta}=[T,\tau_x,\tau_y,\tau_z]^\top$ is linearly related to the control input $\bm{\eta}=B_0\B{u}$, where $\B{u}=[n_1^2,n_2^2,\ldots,n_k^2]^\top$ is the squared motor speeds for a vehicle with $k$ rotors and $B_0$ is the actuation matrix.
The key difficulty stems from disturbance forces $\B{f}_a=[f_{a,x}, f_{a,y}, f_{a,z}]^\top$ and disturbance torques $\bm{\tau}_a=[\tau_{a,x}, \tau_{a,y}, \tau_{a,z}]^\top$, generated by other multirotors. 

\subsection{Swarm Dynamics}
Consider $n$ homogeneous multirotors. To simplify notations, we use $\B{x}^{(i)}=[\B{p}^{(i)};\B{v}^{(i)};R^{(i)};\bm{\omega}^{(i)}]$ to denote the state of the $i^{\text{th}}$ multirotor. Then \cref{eq:dynamics} can be simplified as:
\begin{equation}
\dot{\B{x}}^{(i)} = f(\B{x}^{(i)},\B{u}^{(i)}) + \begin{bmatrix}
\B{0} \\ \B{f}_{a}^{(i)} \\ \B{0} \\ \bm{\tau}_{a}^{(i)}
\end{bmatrix},\ \ i=1,\cdots,n,
\label{eq:onerobot}
\end{equation}
where $f(\B{x}^{(i)},\B{u}^{(i)})$ is the nominal dynamics and $\B{f}_{a}^{(i)}$ and $\bm{\tau}_{a}^{(i)}$ are unmodeled force and torque from interactions between other multirotors.

We use $\B{x}^{(ij)}$ to denote the relative state component between robot $i$ and $j$, e.g., $\B{x}^{(ij)}=[\B{p}^{(j)}-\B{p}^{(i)};\B{v}^{(j)}-\B{v}^{(i)}]$. For robot $i$, the unmodeled force and torque in \cref{eq:onerobot} are functions of relative states to its neighbors,
\begin{equation}
\B{f}_{a}^{(i)} = \B{f}_{a}(\mathcal{N}^{(i)}) \text{ and } \bm{\tau}_{a}^{(i)} = \bm{\tau}_{a}(\mathcal{N}^{(i)}),
\label{eq:interactions}
\end{equation}
where $\mathcal{N}^{(i)}=\{\B{x}^{(ij)}|j\in\text{neighbor}(i)\}$ is the set of the relative states of the neighbors of $i$.
Note that here we assume the swarm system is homogeneous, i.e., each robot has the same functions $f$, $\B{f}_a$, and $\bm{\tau}_a$.

\subsection{Problem Statement \& Approach}
We aim to improve the control performance of a multirotor swarm during close formation flight, by learning the unknown interaction terms $\B{f}_a$ and $\bm{\tau}_a$.
Here, we focus on the position dynamics \cref{eq:pos_dynamics} so $\B{f}_a$ is our primary concern.

We first approximate $\B{f}_a$ using a permutation invariant deep neural network (DNN), and then incorporate the DNN in our exponentially-stabilizing controller.
Training is done offline, and the learned interaction dynamics model is applied in the on-board controller in real-time.
\label{sec:prelim}

\section{Learning Approach}
\label{sec:learning}
We employ state-of-the-art deep learning methods to capture the unknown (or residual) multi-vehicle interaction effects.  In particular, we require that the deep neural nets (DNNs) have strong Lipschitz properties (for stability analysis), can generalize well to new test cases, and use compact encodings to achieve high computational and statistical efficiency.  To that end, we employ deep sets \cite{deepsets} and spectral normalization \cite{bartlett2017spectrally} in conjunction with a standard feed-forward neural architecture.\footnote{An alternative approach is to discretize the input space and employ convolutional neural networks (CNNs), which also yields a permutation-invariant encoding. However, CNNs suffer from two limitations: 1) they require much more training data and computation; and 2) they are restricted to a pre-determined resolution and input domain.} 

\subsection{Permutation-Invariant Neural Networks}
The {permutation-invariant} aspect of the interaction term $\B{f}_a$  \cref{eq:interactions} can be characterized as:
\begin{equation}
\B{f}_a(\mathcal{N}^{(i)}) = \B{f}_a(\pi(\mathcal{N}^{(i)})),
\end{equation}
for any permutation $\pi$.
Since our goal is to learn the function $\B{f}_a$ using DNNs, we need to guarantee that the learned DNN is permutation-invariant.
The following lemma (a corollary of Theorem 7 in~\cite{deepsets}) gives the necessary and sufficient condition for a DNN to be permutation-invariant.
\begin{lemma}[adapted from Thm 7 in~\cite{deepsets}]
A continuous function $h(\{\B{x}_1,\cdots,\B{x}_k\})$, with $\B{x}_i\in[\B{x}_\mathrm{min},\B{x}_\mathrm{max}]^n$, is permutation-invariant if and only if it is decomposable into $\rho(\sum_{i=1}^k\phi(\B{x}_i))$, for  some functions $\phi$ and $\rho$.
\label{lemma:deepsets}
\end{lemma}
%%%%%%%%%%%%%
%\begin{proof}

The proof from \cite{deepsets} is highly non-trivial and only holds for a fixed number of vehicles $k$.  Furthermore, their proof technique (which is likely loose) involves a large expansion in the intrinsic dimensionality (specifically $\rho$) compared to the dimensionality of $h$. We will show in our experiments that  $\rho$ and $\phi$ can be learned using relatively compact DNNs, and can generalize well to larger swarms.

%The sufficient side of Lemma~\ref{lemma:deepsets} is straightforward, since the summation operation is permutation-invariant. The necessary part is not trivial and is proved for fixed $k$ in~\cite{deepsets}.
%\end{proof}

Lemma~\ref{lemma:deepsets} implies we can consider the following ``deep sets'' \cite{deepsets} architecture to approximate $\B{f}_{a}$:
\begin{equation}
\B{f}_{a}^{(i)}=\B{f}_{a}(\mathcal{N}^{(i)})\approx
\rho\left(\sum_{\B{x}^{(ij)}\in \mathcal{N}^{(i)}}\phi(\B{x}^{(ij)},\bm{\theta}_{\phi}),\bm{\theta}_{\rho}\right)=\hat{\B{f}}_a^{(i)},
\label{eq:learningmodel}
\end{equation}
where $\phi(\cdot)$ and $\rho(\cdot)$ are two DNNs, and $\bm{\theta}_{\phi}$ and $\bm{\theta}_{\rho}$ are their corresponding parameters. The output of $\phi$ is a hidden state to represent ``contributions'' from each neighbor, and $\rho$ is a nonlinear mapping from the summation of these hidden states to the total effect. The advantages of this approach are:
\begin{itemize}
\item \textbf{Representation ability.} Since Lemma~\ref{lemma:deepsets} is necessary and sufficient, we do not lose approximation power by using this constrained framework.  We demonstrate strong empirical performance using relatively compact DNNs for $\rho$ and $\phi$.
%Note that the necessary side is only proved to be true for a swarm of fixed size but the sufficient side is always true.

\item \textbf{Computational and sampling efficiency and scalability.} Since the input dimension of $\phi(\cdot)$ is always the same as the single vehicle case, the feed-forward computational complexity of $\cref{eq:learningmodel}$ grows linearly with the number of neighboring vehicles. Moreover, given training data from $n$ vehicles, under the homogeneous dynamics assumption, we can reuse the data $n$ times. In practice, we found that a few minutes flight data is sufficient to accurately learn interactions between two to five multirotors.

\item \textbf{Generalization to varying swarm size.} Given learned $\phi(\cdot)$ and $\rho(\cdot)$, \cref{eq:learningmodel} can be used to predict interactions for any swarm size. In other words, a model trained on swarms of a certain size may also accurately model (slightly) larger swarms. In practice, we found that trained with data from three multirotor swarms, our model can give good predictions for five multirotor swarms.
\end{itemize}

\subsection{Spectral Normalization for Robustness and Generalization}
\label{sec:spectral}
To improve robustness and generalization of DNNs, we use spectral normalization~\cite{bartlett2017spectrally} for training optimization. Spectral normalization stabilizes DNN training by constraining its Lipschitz constant. Spectrally normalized DNNs have been shown to generalize well, which is an indication of stability in machine learning.  Spectrally normalized DNNs have also been shown to be robust, which can be used to provide control-theoretic stability guarantees~\cite{liu2019robust,shi2019neural-lander}. 

Mathematically, the Lipschitz constant of a function $\|g\|_{\text{Lip}}$ is defined as the smallest value such that:
\[\forall \, \B{x}, \B{x}':\ \|g(\B{x})-g(\B{x}')\|_2/\|\B{x}-\B{x}'\|_2\leq \|g\|_{\text{Lip}}.\]
Let $g(\B{x},\bm{\theta})$ be a ReLU DNN parameterized by the DNN weights $\bm{\theta}={W^1,\cdots,W^{L+1}}$:
\begin{equation}
g(\B{x},\bm{\theta}) = W^{L+1}\sigma(W^L\sigma(\cdots \sigma(W^1\B{x})\cdots)),
\end{equation}
where the activation function $\sigma(\cdot)=\max(\cdot,0)$ is called the element-wise ReLU function.
In practice, we apply the spectral normalization to the weight matrices in each layer after each batch gradient descent as follows:
\begin{equation}
W^i \leftarrow W^i / \|W^i\|_2 \cdot \gamma^{\frac{1}{L+1}},i=1,\cdots,L+1,
\label{eq:sn}
\end{equation}
where $\|W^i\|_2$ is the maximum singular value of $W^i$ and $\gamma$ is a hyperparameter. With \cref{eq:sn}, $\|g\|_{\text{Lip}}$ will be upper bounded by $\gamma$. Since spectrally normalized $g$ is $\gamma-$Lipschitz continuous, it is robust to noise $\Delta\B{x}$, i.e., $\|g(\B{x}+\Delta\B{x})-g(\B{x})\|_2$ is always bounded by $\gamma\|\Delta\B{x}\|_2$. In this paper, we apply the spectral normalization on both the $\phi(\cdot)$ and $\rho(\cdot)$ DNNs in \cref{eq:learningmodel}.

\subsection{Data Collection}
Learning a DNN to approximate $\B{f}_a$ requires collecting close formation flight data.
However, the downwash effect causes the nominally controlled multirotors (without compensation for the interaction forces) to move apart from each other, see \cref{fig:intro}.
Thus, we use a cumulative/curriculum learning approach: first, we collect data for two multirotors without a DNN and learn a model.
Second, we repeat the data collection using our learned model as feed-forward term, which allows closer-proximity flight of the two vehicle.
Third, we repeat the procedure with increasing number of vehicles, using the current best model.

Note that our data collection and learning are independent of the controller used and independent of the $\B{f}_a$ compensation.
In particular, if we actively compensate for a learned $\B{f}_a$, this will only affect $\B{f}_u$ in \eqref{eq:pos_dynamics} and not the observed $\B{f}_a$.

\section{Nonlinear Decentralized Controller Design}
\label{sec:theory}
Our \ns{} controller is a nonlinear feedback linearization controller using the learned interaction term $\hat{\B{f}}_a^{(i)}$. Note that \ns{} is decentralized, since $\hat{\B{f}}_a^{(i)}$ is a function of the neighbor set, $\mathcal{N}^{(i)}$, of vehicle $i$. Moreover, the computational complexity of $\hat{\B{f}}_a^{(i)}(\mathcal{N}^{(i)})$ grows linearly as the size of $\mathcal{N}^{(i)}$, since we employ deep sets to encode $\hat{\B{f}}_a^{(i)}$.

\subsection{Reference Trajectory Tracking}
Similar to~\cite{shi2019neural-lander}, we employ an integral controller that accounts for the predicted residual dynamics, which in our case are the multi-vehicle interaction effects. For vehicle $i$, we define the position tracking error as $\Tilde{\B{p}}^{(i)}=\B{p}^{(i)}-\B{p}_d^{(i)}$ and the composite variable $\B{s}^{(i)}$ as:
\begin{equation}
\B{s}^{(i)} = \dot{\Tilde{\B{p}}}^{(i)} + 2\Lambda\Tilde{\B{p}}^{(i)} + \Lambda^2\int\Tilde{\B{p}}^{(i)}dt = \dot{\B{p}}^{(i)} - \B{v}_r^{(i)},
\end{equation}
where $\B{v}_r^{(i)}=\dot{\B{p}}_d^{(i)}-2\Lambda\Tilde{\B{p}}^{(i)}-\Lambda^2\int\Tilde{\B{p}}^{(i)}dt$ is the reference velocity. 
% Note that $\B{s}^{(i)}=0$ is a manifold on which $\Tilde{\B{p}}^{(i)}\rightarrow 0$ exponentially.
We design the total desired rotor force $\B{f}_d^{(i)}$ as:
\begin{equation}
\begin{aligned}
\B{f}_d^{(i)} &= m\dot{\B{v}}_r^{(i)} - K\B{s}^{(i)} - m\B{g} - \hat{\B{f}}_a^{(i)}, \\
\hat{\B{f}}_a^{(i)} &= \rho\Bigl(\sum_{\B{x}^{(ij)}\in \mathcal{N}^{(i)}}\phi(\B{x}^{(ij)},\bm{\theta}_{\phi}),\bm{\theta}_{\rho}\Bigr).
\end{aligned}
\label{eq:position-controller}
\end{equation}
Note that the position control law in \cref{eq:position-controller} is decentralized, because we only consider the relative states $\B{x}^{(ij)}\in \mathcal{N}^{(i)}$ in the controller.

Using $\B{f}_d^{(i)}$, the desired total thrust $T_d^{(i)}$ and desired attitude $R_d^{(i)}$ can be easily computed~\cite{bandyopadhyay2016nonlinear}.
Given $R_d^{(i)}$, we can use any attitude controller to compute $\bm{\tau}_d^{(i)}$, for example robust nonlinear tracking control with global exponential stability~\cite{bandyopadhyay2016nonlinear}, or geometric tracking control on $\mathrm{SE}(3)$~\cite{lee2010geometric}. 
% we can use the following nonlinear attitude controller \todo{change to Lee controller} to track $R_d^{(i)}(t)$~\cite{morgan2016swarm}:
% \begin{equation}
% \begin{aligned}
% \bm{\tau}_d^{(i)} &= J\dot{\bm{\omega}}_r^{(i)} - J\bm{\omega}^{(i)}\times\bm{\omega}_r^{(i)} - K_\omega(\bm{\omega}^{(i)}-\bm{\omega}_r^{(i)}) \\
% &\quad-K_I\int(\bm{\omega}^{(i)}-\bm{\omega}_r^{(i)})dt,
% \end{aligned}
% \label{eq:attitude-controller}
% \end{equation}
% where $\bm{\omega}_r^{(i)}$ is the reference angular velocity. Note that the attitude controller \cref{eq:attitude-controller} is proved to be globally exponentially stable.
From this process, we get $\bm{\eta}_d^{(i)}=[T_d^{(i)};\bm{\tau}_d^{(i)}]$, and then the desired control signal of each vehicle is $\B{u}_d^{(i)}=B_0^{\dag}\bm{\eta}_d^{(i)}$, which can be computed in a decentralized manner for each vehicle.

\subsection{Nonlinear Stability and Robustness Analysis}
Note that since $\|\hat{\B{f}}_a^{(i)}-\B{f}_a^{(i)}\|\neq0$, we can not guarantee the tracking error $\Tilde{\B{p}}^{(i)}\rightarrow 0$. However, under some mild assumptions, we can guarantee input-to-state stability (ISS) using exponential stability~\cite{chung2013phase} for all the vehicles. 
\begin{assumption}
The desired position trajectory $\B{p}_d^{(i)},\dot{\B{p}}_d^{(i)}$, and $\ddot{\B{p}}_d^{(i)}$ are bounded for all $i$. 
\end{assumption}
\begin{assumption}
Define the learning error as $\hat{\B{f}}_a^{(i)}-\B{f}_a^{(i)}$, with two components:  $\hat{\B{f}}_a^{(i)}-\B{f}_a^{(i)}=\bm{\epsilon}^{(i)}_0+\bm{\epsilon}^{(i)}(t)$, where $\bm{\epsilon}^{(i)}_0$ is some constant bias and $\bm{\epsilon}^{(i)}(t)$ is a time-varying term. We assume that for vehicle $i$, $\|\dot{\bm{\epsilon}}^{(i)}(t)\|$ is upper bounded by $d_m^{(i)}$.
\end{assumption}
\begin{theorem}
\label{thm:bound}
Under Assumptions 1 and 2, for vehicle $i$, for some desired trajectory $\B{p}_d^{(i)}(t)$, \cref{eq:position-controller} achieves exponential convergence of the tracking error to an error ball:
\begin{equation}
\lim_{t\rightarrow\infty}\|\Tilde{\B{p}}^{(i)}(t)\|=\frac{d_m^{(i)}}{\lambda_{\min}^2(\Lambda)\lambda_{\min}(K)}.
\label{eq:errorball}
\end{equation}
\end{theorem}
\medskip
\begin{proof}
For vehicle $i$, consider the Lyapunov function $V^{(i)}=\frac{1}{2}m\|\dot{\B{s}}^{(i)}\|^2$. With controller \cref{eq:position-controller}, we get the time-derivative of $V$:
\begin{equation*}
\dot{V}^{(i)}=\dot{\B{s}}^{(i)\top}(-K\dot{\B{s}}^{(i)}+\dot{\bm{\epsilon}}^{(i)}) 
\leq-\dot{\B{s}}^{(i)\top}K\dot{\B{s}}^{(i)}+\|\dot{\B{s}}^{(i)}\|d_m^{(i)}.
\end{equation*}
Using $V^{(i)}=\frac{1}{2}m\|\dot{\B{s}}^{(i)}\|^2$, we have
\begin{equation}
\dot{V}^{(i)}\leq-\lambda_{\min}(K)\frac{2V^{(i)}}{m}+\sqrt{\frac{2V^{(i)}}{m}}d_m^{(i)}.
\end{equation}
Using the Comparison Lemma~\cite{khalil2002nonlinear}, we obtain
\begin{equation*}
\begin{aligned}
\left\lVert\dot{\B{s}}^{(i)}(t)\right\rVert
&\leq \left\lVert\dot{\B{s}}^{(i)}(0)\right\rVert \exp\left(-\frac{\lambda_{\min}(K)}{m}\cdot t\right)+\frac{d_m^{(i)}}{\lambda_{\min}(K)}.
\end{aligned}
\end{equation*}
Note that $\dot{\B{s}}^{(i)}=\ddot{\Tilde{\B{p}}}^{(i)}+2\Lambda\dot{\Tilde{\B{p}}}^{(i)}+\Lambda^2\Tilde{\B{p}}^{(i)}$, and the hierarchical combination between $\dot{\B{s}}^{(i)}$ and $\Tilde{\B{p}}^{(i)}$ results in $\lim_{t\rightarrow\infty}\|\Tilde{\B{p}}^{(i)}\|=\lim_{t\rightarrow\infty}\|\dot{\B{s}}^{(i)}\|/\lambda^2_{\min}(\Lambda)$, yielding \cref{eq:errorball}.
\end{proof}

\section{Experiments}
\label{sec:exp}
We use a slightly modified Crazyflie 2.0 (CF) as our quadrotor platform, a small (\SI{9}{cm} rotor-to-rotor) and lightweight (\SI{34}{g}) product that is commercially available.
We use the Crazyswarm~\cite{crazyswarm} package to control multiple Crazyflies simultaneously.
Each quadrotor is equipped with four reflective markers for pose tracking at \SI{100}{Hz} using a motion capture system.
The nonlinear controller, extended Kalman filter, and neural network evaluation are running on-board the STM32 microcontroller.

For data collection, we use the uSD card extension board and store binary encoded data roughly every \SI{10}{ms}.
Each dataset is timestamped using the on-board microsecond timer and the clocks are synchronized before takeoff using broadcast radio packets.
The drift of the clocks of different Crazyflies can be ignored for our short flight times (less than \SI{2}{min}).

\subsection{Calibration and System Identification}
\label{sec:systemid}
Prior to learning the residual term $\B{f}_a$, we first calibrate the nominal dynamics model $f(\B{x},\B{u})$.
We found that existing motor thrust models \cite{BitcrazeThrust, systemid} are not very accurate, because they only consider a single motor and ignore the effect of the battery state of charge.
We calibrate each Crazyflie by mounting the whole quadrotor on a \SI{100}{g} load cell which is directly connected to a custom extension board.
We collect the current battery voltage, PWM signals (identical for all 4 motors), and measured force from the load cell for various motor speeds.
We use this data to find two polynomial functions.  The first  computes the PWM signal given the current battery voltage and desired force.  The second computes the maximum achievable force, given the current battery voltage.
This second function is important for thrust mixing when motors are saturated~\cite{DBLP:journals/ral/FaesslerFS17}.

We notice that the default motors and propellers can only produce a total force of about \SI{48}{g} with a full battery, resulting in a best-case thrust-to-weight ratio of 1.4.
Thus, we replaced the motors with more powerful ones (that have the same physical dimensions) to improve the best-case thrust-to-weight ratio to 2.6.
We use the remaining parameters ($J$, thrust-to-torque ratio) from the existing literature~\cite{systemid}.

\subsection{Data Collection and Learning}
We utilize two types data collection tasks: random walk and swapping.
For random walk, we implement a simple reactive collision avoidance approach based on artificial potentials on-board each Crazyflie~\cite{artificialPotentials}.
The host computer randomly selects new goal points within a small cube for each vehicle in a fixed frequency. Those goal points are used as an attractive force, while neighboring vehicles contribute a repulsive force.
For swapping, we place vehicles in different horizontal planes on a cylinder and let them move to the opposite side. All vehicles are vertically aligned for one time instance, causing a large interaction force, see \cref{fig:intro,fig:swaps,fig:validation} for examples with two, three, and four vehicles.
The random walk data helps us to explore the whole space quickly, while the swapping data ensures that we have data for a specific task of interest. For both task types, we varied the scenarios from two to four vehicles, and collected one minute of data for each scenario. %, we collected only one minute swapping data and one minute random walk data for each case.

To learn the interaction function $\B{f}_a(\mathcal{N}^{(i)})$, we collect the timestamped states $\B{x}^{(i)}=[\B{p}^{(i)};\B{v}^{(i)};\dot{\B{v}}^{(i)};R^{(i)};\B{f}_u^{(i)}]$ for each vehicle $i$.
We then compute $\B{y}^{(i)}$ as the observed value of $\B{f}_a(\mathcal{N}^{(i)})$. We compute $\B{f}_a^{(i)}=\B{f}_a(\mathcal{N}^{(i)})$  using $m\dot{\B{v}}^{(i)}=m\B{g}+R^{(i)}\B{f}_u^{(i)}+\B{f}_a^{(i)}$ in \cref{eq:pos_dynamics}, where $\B{f}_u^{(i)}$ is calculated based on our system identification in \cref{sec:systemid}.
Our training data consists of sequences of $(\mathcal{N}^{(i)},\B{y}^{(i)})$ pairs, where $\mathcal{N}^{(i)}=\{\B{x}^{(ij)}|j\in\text{neighbor}(i)\}$ is the set of the relative states of the neighbors of $i$.
In practice, we compute the relative states from our collected data as $\B{x}^{(ij)}=[\B{p}^{(j)}-\B{p}^{(i)};\B{v}^{(j)}-\B{v}^{(i)}]\in\mathbb{R}^6$ (i.e.,  relative global position and relative global velocity), since the attitude information $R$ and $\bm{\omega}$ are not dominant for $\B{f}_a$. In this work, we only learn the $z$ component of $\B{f}_a$ since we found the other two components, $x$ and $y$, are very small, and do not significantly alter the nominal dynamics.

Since our swarm is homogeneous, each vehicle has the same function $\B{f}_a$.
Thus, we stack all the vehicle's data and train on them together, which implies more training data overall for larger swarms. Let $\mathcal{D}^{(i)}$ denote the training data of vehicle $i$, where the input-output pair is $(\mathcal{N}^{(i)},\B{y}^{(i)})$. We use the ReLU network class for both $\phi$ and $\rho$ neural networks and our training loss is:
\begin{equation}
\begin{aligned}
\sum_i\,\sum_{\mathcal{D}^{(i)}}\,
\Bigl\rVert\rho\Bigl(\sum_{\B{x}^{(ij)}\in \mathcal{N}^{(i)}}\phi(\B{x}^{(ij)},\bm{\theta}_{\phi}),\bm{\theta}_{\rho}\Bigr)-\B{y}^{(i)}\Bigr\rVert_2^2,
\end{aligned}
\end{equation}
where $\bm{\theta}_{\phi}$ and $\bm{\theta}_{\rho}$ are neural network weights to be learned. Our $\phi$ DNN has four layers with architecture $6\rightarrow25\rightarrow40\rightarrow40\rightarrow40$, and our $\rho$ DNN also has four layers, with architecture $40\rightarrow40\rightarrow40\rightarrow40\rightarrow1$.
%Note that the output of $\phi$ is a hidden state to represent ``contributions'' from each neighbor, and the input to $\rho$ is the summation of all these hidden states. In our work, the dimension of the hidden state is $40$.
We use PyTorch~\cite{pyTorch} for training and implementation of spectral normalization (see \cref{sec:spectral}) of $\phi$ and $\rho$. 
We found that spectral normalization is in particular important for the small Crazyflie quadrotors, because their IMUs are directly mounted on the PCB frame causing more noisy measurements compared to bigger quadrotors.

Using the learned weights $\bm{\theta}_{\phi}$ and $\bm{\theta}_{\rho}$, we generate C-code to evaluate both networks efficiently on-board the quadrotor, similar to prior work~\cite{sim-to-multi-real}.
The STM32 \SI{168}{MHz} microcontroller can evaluate each of the networks in about \SI{550}{\micro\second}.
Thus, we can compute $\B{f}_a$ in less than \SI{4}{ms} for 6 or less neighbors, which is sufficient for real-time operations.

\subsection{Neural-Swarm Control Performance}

\begin{table}
\caption{Maximum $z$-error (in meters) for varying swarm size swapping tasks and neural networks. Training on more vehicles leads to the best overall performance for all swarm sizes.}
\label{tab:nnperf}
\centering
\begin{tabular}{c||p{0.7cm}|p{0.7cm}|p{0.7cm}|p{0.7cm}}
\diagbox{Controller}{Flight test} & 2 CF Swap & 3 CF Swap & 4 CF Swap & 5 CF Swap\\
\hline
\hline
Baseline & 0.094 & 0.139 & 0.209 & 0.314 \\
Trained w/ 2 CF & 0.027 & 0.150 & 0.294 & N.A.\\
Trained w/ 3 CF & 0.026 & 0.082 & 0.140 & 0.159\\
Trained w/ 4 CF & 0.024 & 0.061 & 0.102 & 0.150\\
\end{tabular}
\vspace{-15pt}
\end{table}

\begin{figure}
\includegraphics[width=\linewidth]{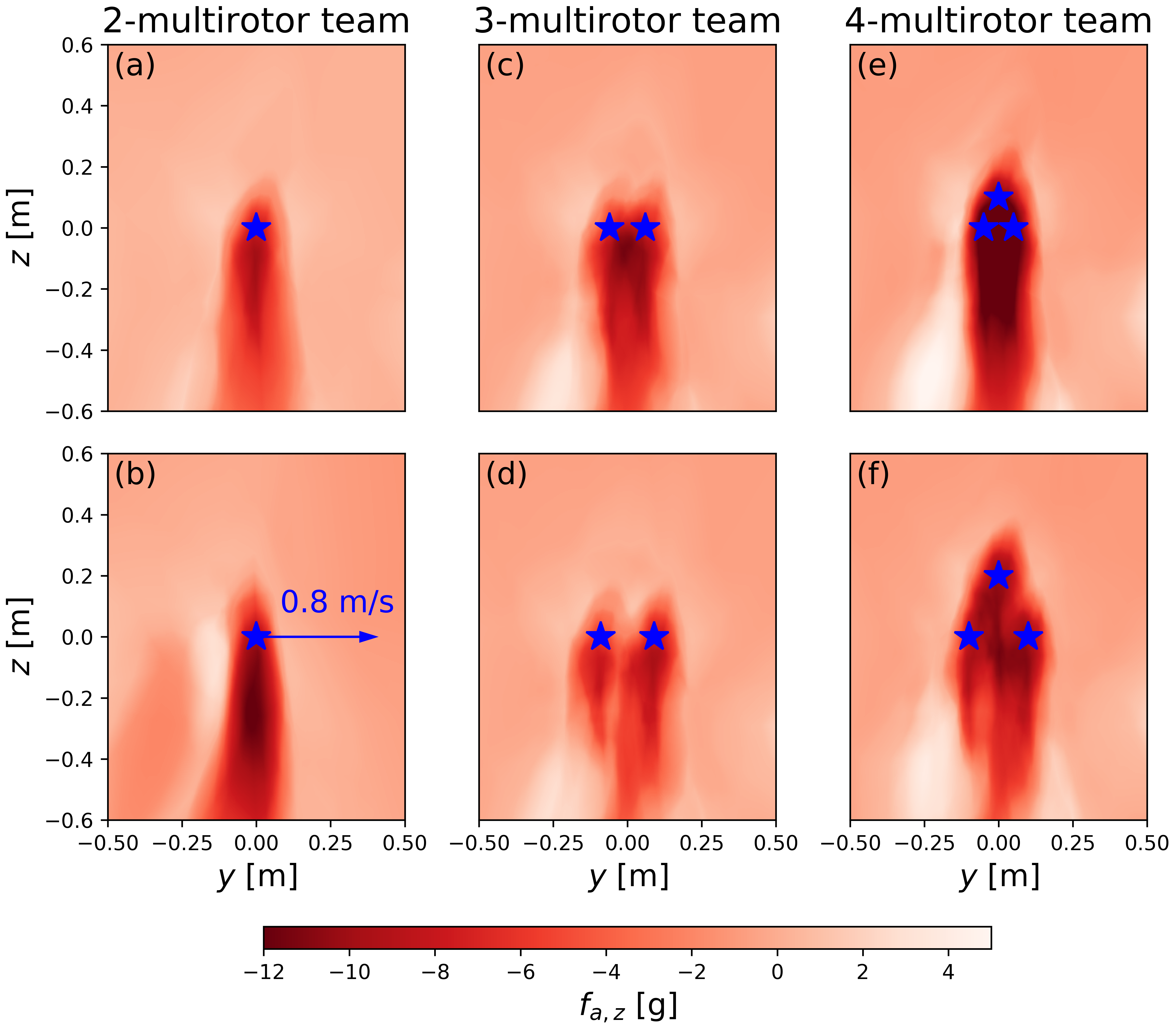}
\caption{$\hat{f}_{a,z}$ generated by $\phi$ and $\rho$ networks trained with 3 CF data. Each heatmap gives the prediction of $f_{a,z}$ of a vehicle in different horizontal and vertical (global) positions, and the (global) position of neighboring vehicles are represented by blue stars. A more detailed explanation is in \cref{sec:nnvis}.}
\label{fig:heatmap}
\vspace{-15pt}
\end{figure}

We study the performance and generalization of different controllers on a swapping task using varying number of quadrotors.
An example of our swapping task for two vehicles is shown in \cref{fig:intro}.
The swapping task for multiple vehicles causes them to align vertically at one point in time with vertical distances of \SIrange{0.2}{0.25}{m} between neighbors.
This task is challenging, because: i) the lower vehicles experience downwash from multiple vehicles flying above;  ii) the different velocity vectors of each vehicle creates interesting effects, including an effect where $\B{f}_a$ is positive for a short period of time (see \cref{fig:heatmap}(b) for an example); and iii) for the case with more than two vehicles, the aerodynamic effect is not a simple superposition of each pair (see \cref{fig:heatmap}(c-f) for examples).
%\todo{does this effect have a name?}

We use the following four controllers: 1) The baseline controller uses our position tracking controller \cref{eq:position-controller} with $\hat{\B{f}}_a^{(i)}\equiv0,\forall i$ and a nonlinear attitude tracking controller~\cite{lee2010geometric}; 2) -- 4) The same controller with the same gains, but $\hat{\B{f}}_a^{(i)}$ computed using different neural networks (trained on data flying 2, 3, and 4 quadrotors, respectively.)
Note that all controllers, including the baseline controller, always have integral control compensation parts.
Though an integral gain can cancel steady-state error during set-point regulation, it can struggle with complex time-variant interactions between vehicles. This issue is also reflected in the tracking error bound in \cref{thm:bound}. In \cref{thm:bound}, the tracking error will converge to $d_m^{(i)}/\lambda_{\min}^2(\Lambda)\lambda_{\min}(K)$. For our baseline we have $d_m^{(i)}=\sup_t\|\frac{d}{dt}\B{f}_a^{(i)}(t)\|$, which means if $\B{f}^{(i)}_a$ is changing fast as in the swapping task, our baseline will not perform well.  

\begin{figure}
%\vspace{-0.3in}
\includegraphics[width=\linewidth]{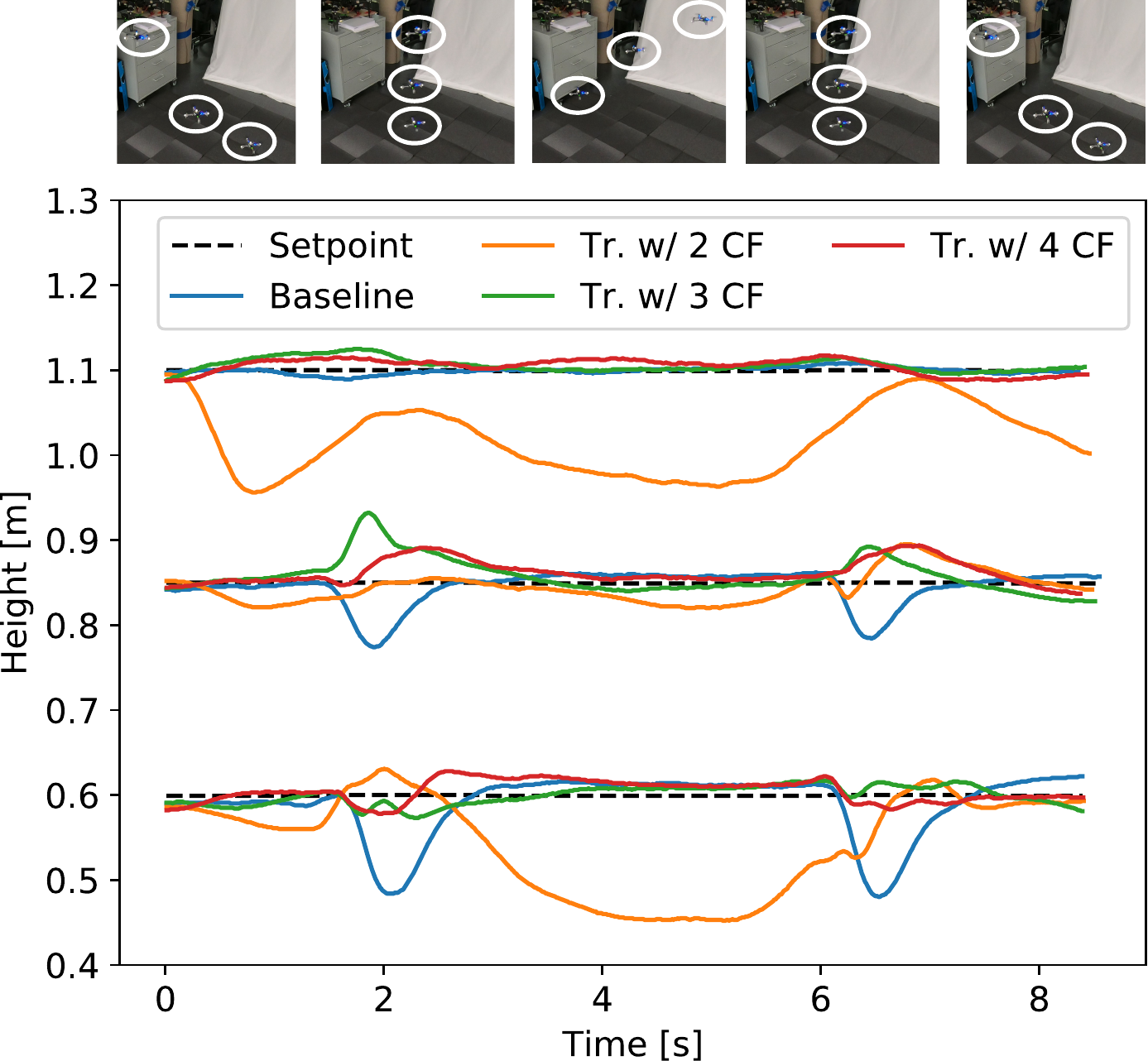}
%\vspace{-0.1in}
\caption{Three vehicles moving using different control policies (corresponding to \cref{tab:nnperf}, column 2). Each quadrotor flies at a different fixed height (\SI{25}{cm} vertical separation) and swaps sides such that all vehicles align vertically at $t=\SI{2}{s}$ and $t=\SI{6.5}{s}$. Our approach trained on 3 or 4 vehicles controls the height much better than the baseline approach.}
\label{fig:swaps}
\vspace{-15pt}
\end{figure}

We repeat the swapping task for each controller six times, and report the maximum $z$-error that occurred for any vehicle over the whole flight.
We also verified that the $x$- and $y$-error distributions are similar across the different controllers and do not report those numbers for brevity.

\textbf{Results.} Our results, described in \cref{tab:nnperf}, show three important results: i) our controller successfully reduces the worst-case $z$-error by a factor of two to four (e.g., \SI{2.4}{cm} instead of \SI{9.4}{cm} for the two vehicle case); ii) our controller successfully generalizes to cases with more vehicles when trained with at least three vehicles (e.g., the controller trained with three quadrotors significantly improves flight performance even when flying five quadrotors); and iii) our controllers do not marginalize small-vehicle cases (e.g., the controller trained with four quadrotors works very well for the two-vehicle case).
The observed maximum $z$-error for the test cases with three to five quadrotors is larger compared to the two-vehicle case because we occasionally saturate the motors during flight.

\cref{fig:swaps} depicts an example of the swapping task for three quadrotors (showing two out of the six swaps), which corresponds to column 2 of \cref{tab:nnperf}. We observe that: 
i) when trained on at least three quadrotors, our approach significantly outperforms the baseline controller;
and ii) the performance degrades significantly when only trained on two quadrotors, since the training data does not include data on superpositions.
%Examples of the flight performance on swapping task for 3 quadrotors (showing 2 out of the 6 swaps) is shown in \cref{fig:swaps}. \cref{fig:swaps} presents performances under four different controllers: baseline, and \ns{} with three different neural networks, trained on flying data of 2, 3 and 4 quadrotors, respectively. In other word, \cref{fig:swaps} is corresponding to the column 2 of \cref{tab:nnperf}.

\subsection{Learned Neural Network Visualization}
\label{sec:nnvis}
\cref{fig:heatmap} depicts the prediction of $f_{a,z}$, trained with flying data of 3 multirotors.
The color encodes the magnitude of $\hat{f}_{a,z}$ for a single multirotor positioned at different global $(y,z)$ coordinates. 
The blue stars indicate the (global) coordinates of neighboring multirotors. All quadrotors are in the same $x$-plane.
For example, in \cref{fig:heatmap}(c) there are two quadrotors hovering at $(0,-0.06, 0)$ and $(0,0.06,0)$.
If we place a third quadrotor at $(0,-0.3,-0.5)$, it would estimate $\hat{f}_{a,z}=\SI{4}{g}$ as indicated by the white color in that part of the heatmap.
All quadrotors are assumed to be stationary except for \cref{fig:heatmap}(b), where the one neighbor is moving at \SI{0.8}{m/s}.

We observe that the interaction between quadrotors is non-stationary and sensitive to relative velocity, as well as not a simple superposition between pairs. In \cref{fig:heatmap}(b), the vehicle's neighbor is moving, and the prediction becomes significantly different from \cref{fig:heatmap}(a), where the neighbor is just hovering. Moreover, in \cref{fig:heatmap}(b) there is an interesting region with relatively large positive $\hat{f}_{a,z}$, which is consistent with our observations in flight experiments.
We can also observe that the interactions are not a simple superposition of different pairs. For instance, \cref{fig:heatmap}(e) shows a significantly stronger updraft effect outside the downwash region than expected from a simple superposition of the prediction in \cref{fig:heatmap}(a).
%We also observe that the interactions are not simple superposition of different pairs, see \cref{fig:heatmap}(c-f).  

\begin{figure}
\includegraphics[width=\linewidth]{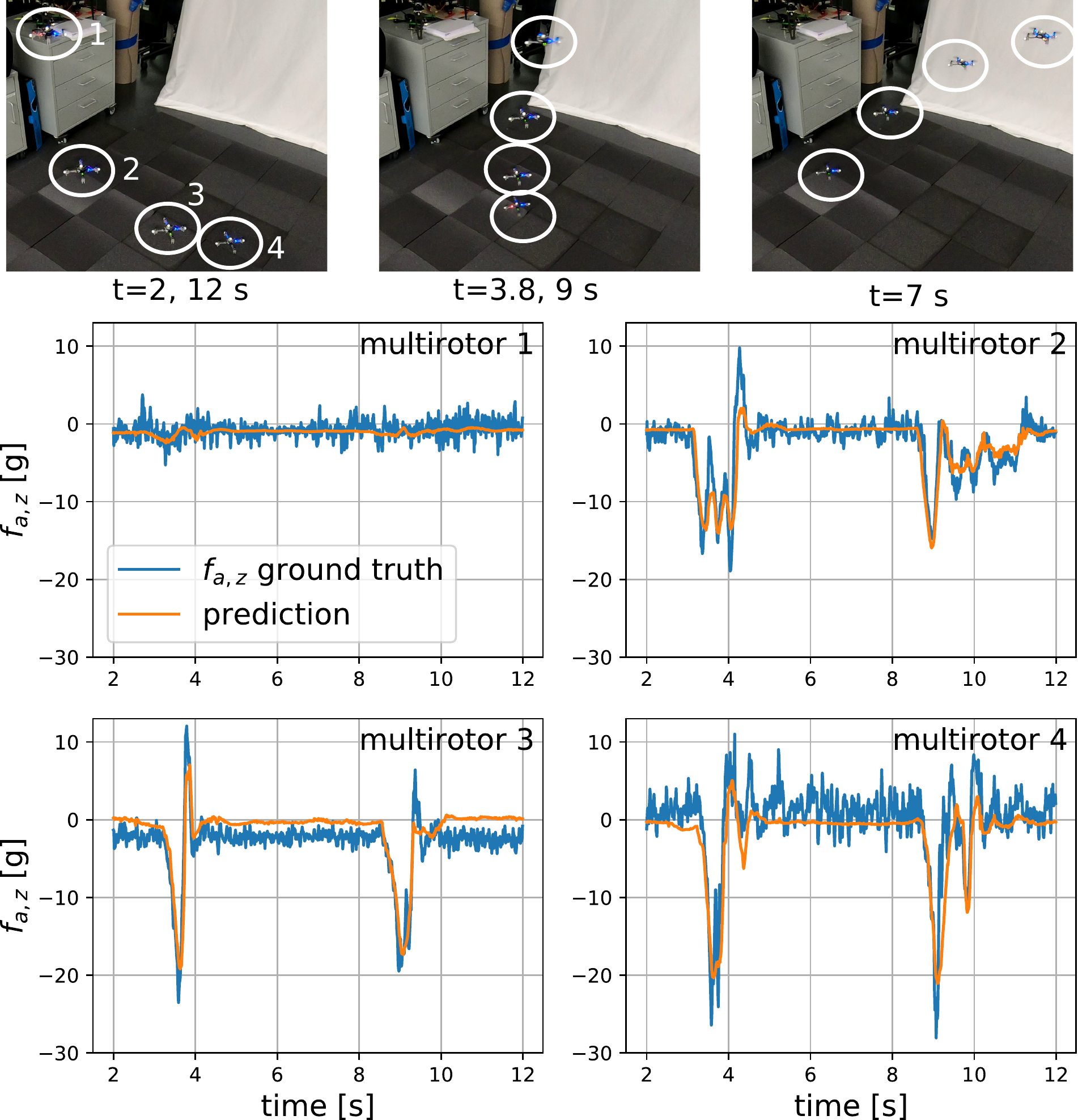}
\caption{Ground truth and $\hat{f}_{a,z}$ for a 4-vehicle swapping task. $\hat{f}_{a,z}$ is generated by neural networks trained with data from three vehicles. Our method generalizes well and predicts $f_{a,z}$ with high accuracy.}
\label{fig:validation}
\vspace{-15pt}
\end{figure}

Our approach can generalize well using data for 3 vehicles to a larger 4-vehicle system. In \cref{fig:heatmap}, all the predictions are from $\phi$ and $\rho$ networks trained with 3 CF data, but predictions for a 4-vehicle team (as shown in \cref{fig:heatmap}(e-f)) are still reasonable and work well in real flight tests (see \cref{tab:nnperf} and \cref{fig:swaps}). 
For this 4 CF swapping task, we compare ground truth $f_{a,z}$ and its prediction in \cref{fig:validation}. As before, the prediction is computed using neural networks trained with 3 CF flying data. We found that 1) for multirotor 3 and 4, $f_{a,z}$ is so high such that we cannot fully compensate it within our thrust limits; and 2) the prediction matches the ground truth very well, even for complex interactions (e.g., multirotor 2 in \cref{fig:validation}), which indicates that our approach generalizes well.

% \subsection{Dense Formation Flight}

% \todo{figure 8/triangle experiments; 10-20 CFs experiment}

\section{Conclusion}
\label{sec:con}
In this paper, we present a decentralized controller that enables close-proximity flight of homogeneous multirotor teams.
Our solution, \ns, uses deep neural networks to learn the interaction forces between multiple quadrotors and only relies on relative positions and velocities of neighboring vehicles.
We demonstrate in flight tests that our training method generalizes well to a varying number of neighbors, is computationally efficient, and reduces the worst-case height error by a factor of two or better.
To our knowledge, our solution is the first that models interactions between more than two multirotors.

There are many directions for future work.  First, one can extend our work to heterogeneous swarms, which may require extending the neural net architecture beyond spectral normalized deep sets.  Second, one can use the learned interaction forces for motion planning and control of dynamically changing formations. 
Third, one can learn $\bm{\tau}_a$ as well as $\B{f}_a$ to improve the flight performance during aggressive maneuvers even further.

%Third, one can study how to efficiently scale up our approach to massive swarm sizes, which poses significant challenges in data collection and computation.

%In the future, we plan to extend our method to heterogeneous teams, use the learned interaction forces for motion planning, and learn $\B{\tau}_a$ for more aggressive flights.

% future work:
% * use for motion planning
% * learn tau_a
% * look at tasks other than swapping (e.g., continuous flight on top of each other)
% * heterogeneous teams

% Comment the following line for arXiv
%\addtolength{\textheight}{-7cm}   % This command serves to balance the column lengths
                                  % on the last page of the document manually. It shortens
                                  % the textheight of the last page by a suitable amount.
                                  % This command does not take effect until the next page
                                  % so it should come on the page before the last. Make
                                  % sure that you do not shorten the textheight too much.

%%%%%%%%%%%%%%%%%%%%%%%%%%%%%%%%%%%%%%%%%%%%%%%%%%%%
% Comment the following line for arXiv
%\newpage

\bibliographystyle{IEEEtran}

\newpage
\balance
\bibliography{IEEEabrv,ref}

% \section{TODO}

% \begin{enumerate}
%     \item (if time) 10-20 CFs upstairs without motor motification (eg.: 3d swap using existing motion planner with spherical collision model); cmp w/ wo/ NN 
%     \item decentralized: mention neighboring range
%     \item add \cite{doi:10.2514/1.7579}
%     \item upload data collection videos
% \end{enumerate}

\end{document}